\documentclass[letterpaper, 10 pt, conference]{ieeeconf}  

\IEEEoverridecommandlockouts                              

\usepackage{graphicx}
\usepackage{xcolor}
\usepackage{booktabs}
\usepackage{pifont}
\usepackage{ragged2e}
\usepackage{adjustbox}
\usepackage{multirow}
\usepackage{color,soul}
\usepackage{url}
\usepackage{hyperref} 
\usepackage{epsfig} 
\usepackage{mathptmx} 
\usepackage{times} 
\usepackage{amsmath} 
\usepackage{amssymb}  

\title{{\bf DISC}:  Dataset for Analyzing\\{\bf D}riving Styles {\bf I}n {\bf S}imulated {\bf C}rashes for Mixed Autonomy}

\author{Sandip Sharan Senthil Kumar, Sandeep Thalapanane,\\Guru Nandhan Appiya Dilipkumar Peethambari, Sourang SriHari, Laura Zheng, and Ming C. Lin \\
    \thanks{The authors are with the Department of Computer Science, 
         University of Maryland at College Park, MD, U.S.A.
        E-mail: \{sandeept, sandip26, guruadp, sourang, lyzheng, lin\}@umd.edu} %
       \href{https://gamma.umd.edu/disc}{\texttt{gamma.umd.edu/disc}}
       }

\begin{document}

\maketitle
\thispagestyle{empty}
\pagestyle{empty}


\begin{figure*}[t]
    \centering
    \includegraphics[width=0.95\textwidth]{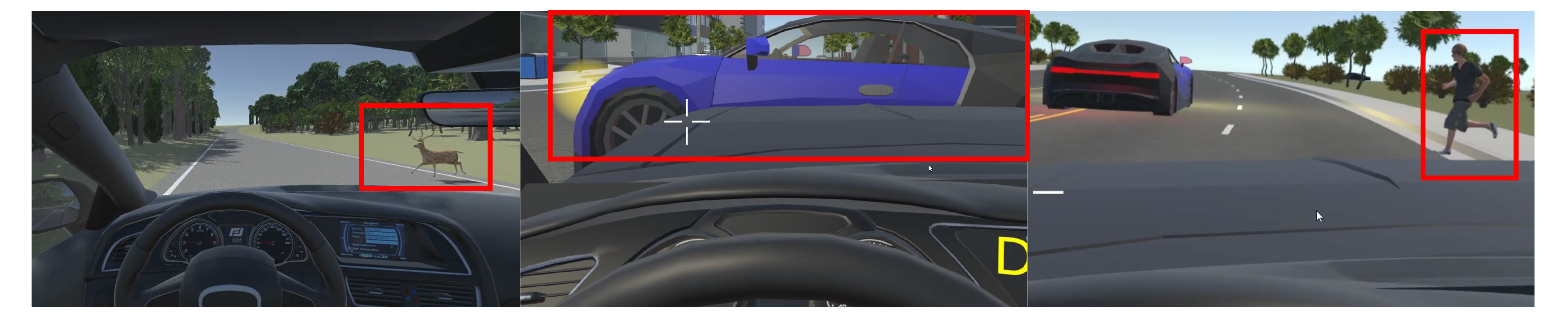}
    \vspace*{-1em}
    \caption{\textbf{Scenarios from the driver's point of view:} The LEFT image depicts the deer-crossing scenario, where the participant encounters a running deer while driving on a country road. The MIDDLE image shows the running red lights scenario, where a participant encounters a car at an intersection that crosses the road after ignoring the red traffic signal. The RIGHT image displays a jaywalking pedestrian from the driver's point of view.}
    \label{fig:sim}
    \vspace*{-1em}
\end{figure*}

\begin{abstract}

Handling pre-crash scenarios is still a major challenge for self-driving cars due to limited practical data and human-driving behavior datasets. We introduce DISC (Driving Styles In Simulated Crashes), one of the first datasets designed to capture various driving styles and behaviors in pre-crash scenarios for mixed autonomy analysis. 
DISC includes over 8 classes of driving styles/behaviors from hundreds of drivers navigating a simulated vehicle through a virtual city, encountering rare-event traffic scenarios. 
This dataset enables the classification of pre-crash human driving behaviors in unsafe conditions, supporting individualized trajectory prediction based on observed driving patterns. 
By utilizing a custom-designed VR-based in-house driving simulator, {\em TRAVERSE}, data was collected through a driver-centric study involving human drivers encountering twelve simulated accident scenarios. 
This dataset fills a critical gap in human-centric driving data for rare events involving interactions with autonomous vehicles. 
It enables autonomous systems to better react to human drivers and optimize trajectory prediction in mixed autonomy environments involving both human-driven and self-driving cars. 
In addition, individual driving behaviors are classified through a set of standardized questionnaires, carefully designed to identify and categorize driving behavior traits. 
We correlate data features with driving behaviors, showing that the simulated environment reflects real-world driving styles. 
DISC is the first dataset to capture how various driving styles respond to accident scenarios, offering significant potential to enhance autonomous vehicle safety and driving behavior analysis in mixed autonomy environments.

\end{abstract}

\section{INTRODUCTION}

Recent advances in self-driving platforms,  from food delivery robots to autonomous vehicles (AV), highlight the urgent need for strong safety standards.
Leading companies in the autonomous vehicle industry have pioneered self-driving taxis capable of navigating complex traffic situations, often achieving safety levels surpassing those of human drivers.

Despite these advancements, real-world data collection, using sensors such as LiDAR and cameras, is often limited to mostly normal driving scenarios and particularly lacking in accidental events. This paper addresses a considerable gap in current research: \textit{the lack of adequate data collected on rare events and accidents}. In order for autonomous vehicles (AVs) to learn to handle adversarial events (e.g. accidents), compounded by unpredictable factors such as jaywalking pedestrians and adverse weather conditions, it remains a major challenge to guarantee safe autonomous driving. 


Current autonomous driving systems are predominantly trained using datasets collected from safe, typical traffic conditions. While this approach ensures robustness for normal driving scenarios, it inadvertently leaves AVs ill-prepared for unexpected and dangerous scenarios. This scarcity of data on adverse driving events results in AVs often not performing optimally when encountered with rare and dangerous scenarios. Addressing this issue requires innovative methods for generating realistic and high-quality training data encompassing a broader spectrum of {\em unsafe} and {\em hazardous} driving conditions. Typically, synthetic risky scenarios are created using simulations, but these simulations frequently lack the realism and accuracy needed to fully capture the complexity of adversarial events.

Furthermore, in the near horizon during the deployment of self-driving vehicles, we can expect the AVs to operate in a `{\bf mixed autonomy}', where there will be human-driven and self-driving cars operating side by side on the road for the foreseeable future.  AVs require the capability to anticipate human-driven vehicle behaviors that will likely be different and less predictable than a self-driving car, thereby making it especially critical to capture large amounts of human driving data under diverse pre-crash scenarios.

This work aims to bridge this gap by employing a virtual reality (VR) vehicle simulator, TRAVERSE\cite{thalapanane2024traversetrafficresponsiveautonomousvehicle}, to immerse participants in various accident scenarios. This approach enables safe collections of human driving data and behaviors in high-risk situations, significantly enhancing our understanding of driving dynamics and safety. TRAVERSE is specifically designed to simulate rare event scenarios based on the National Highway Traffic Safety Administration (NHTSA) pre-crash typology \cite{najm2007definition}, ensuring that the assessed scenarios are both impactful in the real world via quantified societal costs, in addition to being realistic and relevant for participants. 

Participants also complete a driving style test based on the Multidimensional Driving Style Inventory (MDSI) \cite{TAUBMANBENARI2016179} questionnaire to analyze their driving style. This comprehensive questionnaire enables us to classify driving behaviors into eight distinct traits as identified by the MDSI, providing valuable insights into driving styles.


\noindent
This paper presents three key contributions: 

\begin{itemize}
\item The DISC dataset comprises 2,527,004 sets of sensory data for 1,205 vehicle trajectories from 110 participants, capturing diverse driving styles and behaviors in pre-crash scenarios for mixed autonomy analysis, providing a critical foundation for trajectory prediction models.
\item Each collected set of sensory data consists of braking, acceleration, steering, and eye-tracking metrics, collected using Meta Quest Pro, which allows for a deeper analysis of driver perception and responses.
\item DISC consists of diverse data generated under adverse weathers (fog, snow, rain) and varying time/lighting settings (dawn, morning, noon, dusk, night).
\end{itemize}

In summary, our approach utilizes VR-based simulations to produce essential training data for autonomous vehicles in high-risk scenarios. By replicating real-world traffic conditions, we enhance the accuracy of models predicting human-driven vehicle behaviors, thereby improving autonomous vehicle navigation in mixed-autonomy environments. By leveraging the newly introduced human-driving style dataset, {\bf DISC}, we demonstrate significant potential for advancing the safety and efficiency of autonomous driving systems.

\section{Related Works}

\subsection{Driving Simulators}

High-fidelity driving simulators, such as the NADS1 Driving Simulator \cite{chen2001nads, haug1990feasibility, 10179024} and the WTI Simulator \cite{kelly2007high}, offer exceptional realism but face deployment challenges due to their high costs and logistical requirements.

In contrast, low-fidelity simulators present more cost-effective and portable options. The University of Iowa’s miniSim \cite{miniSim} strikes a balance between realism and practicality, while Deep Drive \cite{wijaya2022deepdrive}, is tailored for self-driving AI, utilizing Unreal Engine. CARLA \cite{dosovitskiy2017carla} provides detailed urban environments with dynamic traffic and weather, and LGSVL \cite{rong2020lgsvl} enhances these features with physics-based models and Unity compatibility. However, these simulators often fall short of realism which is vital for behavioral studies.

To address this gap, VR simulators such as DrEyeVR \cite{silvera2022dreyevr}, built on CARLA, offer improved realism but face challenges with customizability and frame rates. Our in-house simulator, TRAVERSE, integrates Unity, SUMO, and Road Runner \cite{gomez2021train} to create dynamic driving scenarios with environmental effects, enhancing the realism for behavioral research.
\vspace{-1mm}

\subsection{Driving Datasets}

\subsubsection{Sensory and Motion Forecasting Datasets}

Recent advancements in autonomous driving datasets have significantly improved scenario realism. The Waymo Open Motion Dataset \cite{ettinger2021large} offers extensive data on perception and motion forecasting in diverse urban settings. The Argoverse-2 Dataset \cite{argoverse} enhances its predecessor with richer sensor data and advanced 3D tracking. nuScenes \cite{9156412} provides varied driving conditions with multiple sensors, while nuPlan \cite{caesar2022nuplanclosedloopmlbasedplanning} focuses on planning algorithm evaluation.

KITTI \cite{Geiger2013IJRR} excels in 3D object detection and visual odometry, whereas the Lyft Level 5 Dataset \cite{houston2020thousandhoursselfdrivingmotion} and Pandaset \cite{xiao2021pandasetadvancedsensorsuite} provide detailed 3D object detection in urban contexts. ApolloScape \cite{wang2019apolloscape} offers 3D scene understanding, and Berkeley DeepDrive \cite{yu2020bdd100kdiversedrivingdataset} addresses diverse driving conditions. The CADC Dataset \cite{Pitropov_2020} and the Oxford Radar RobotCar Dataset \cite{barnes2020oxfordradarrobotcardataset} are critical for testing in adverse weather conditions. Comma2k19 \cite{comma2k19} provides highway driving data for fused pose estimators and mapping algorithms. Despite their strengths, these datasets often focus on safe driving conditions and lack coverage of accident scenarios, which are essential for autonomous vehicle safety.

Studies by Kolla et al. (2022) \cite{KOLLA202294} and Ren et al. (2023) \cite{REN2023107021} propose methods for reconstructing traffic situations and analyzing crash reports, while Zhijun et al. (2021) \cite{s21175767} use GANs to generate traffic accident data. Nonetheless, limitations persist, underscoring the need for datasets including pedestrian, animal, and non-motor vehicle incidents.

\subsubsection{Autonomous Driving and Driving Behavior}

Ruoxuan et al. (2024) \cite{yang2024driving} propose a multi-alignment framework to synchronize LLM-powered drivers with human styles using CARLA simulations. Yi et al. (2021) \cite{RePEc:plo:pone00:0254047} introduce a collaborative driving style classification method with superior simulation performance. However, the reliability of these studies' real-world data is questioned due to potential limitations in driving behavior authenticity.

Our study addresses this by incorporating customizable adversarial events like jaywalking pedestrians and deer crossing within a driving simulator. The DISC dataset fills the gap in human-centric driving data for rare events involving autonomous vehicles, supporting the development of machine-learning models with metrics such as acceleration, braking, vehicle position, eye position, lane data, and road conditions.

\section{DISC Dataset for Mixed Autonomy}

Our primary objective is to develop a diverse set of intricate simulated accident scenarios, drawing inspiration from the NHTSA pre-crash typology. The adoption of the NHTSA typology as the basis for scenario design is motivated by its comprehensive and standardized framework for categorizing real-world accident types. This approach ensures a structured and controlled analysis of driver responses, systematically addressing various aspects of driver decision-making and adaptability in potential collision scenarios. The study comprises twelve distinct scenarios, detailed in Table \ref{ScenarioDescriptions}, alongside a practice scenario aimed at familiarizing users with vehicle controls and the simulation environment. This practice session enhances participant immersion and ensures the robustness of the dataset.

To counteract order bias, participants experience a randomized sequence of scenarios. Each scenario, lasting 30 to 45 seconds, offers a comprehensive evaluation of participant reaction times across a spectrum of accident contexts. To prevent the participants from having increased alertness and overly cautious behavior, a set of non-crash scenarios (Scenario 9-12) is introduced. These scenarios are included to assess driver decision-making in safer contexts while ensuring alignment with the study's objectives.

Human driving data collection occurs within a custom-designed VR-based in-house driving simulator named TRAVERSE, comprising twelve rare event simulations. Participants are asked to maneuver scenarios following route instructions. Breaks are allowed to prevent motion sickness and bias, ensuring data quality. Data collection involves additional hardware components for user engagement and system response insights, including a Logitech steering wheel for vehicle control and MetaVR technology for immersive virtual reality as shown in Figure \ref{fig:sim}. The collected data encompasses driving style data using the MDSI and sensory data from the traffic simulator and game engine, as elaborated in the following sub-sections.

\begin{table}[t]
    \caption{Description of Simulated Scenarios}
    \label{ScenarioDescriptions}
    \centering
    \vspace*{-1em}
    \renewcommand{\arraystretch}{1.25}
    \begin{tabular}{p{0.1cm}p{1.8cm}p{5.2cm}}
        \toprule
        \textbf{No.} & \textbf{Title} & \textbf{Description} \\
        \midrule
        1 & Sudden Lane Intercept & Sudden, unsignaled lane changes that may cause collisions, require a quick response. \\
        
        2 & Crash at T-Bone Intersection & A perpendicular collision with significant damage and injury risks. \\
        3 & Sudden Vehicle Stop & Sudden stops due to obstacles or emergencies, requiring drivers to react quickly. \\

        4 & Running Red Lights & Vehicles running red lights at intersections, increasing collision risks. \\
        5 & Deer Crossing & Unexpected animal crossing, requiring fast decisions to avoid accidents. \\
        6 & Crash at Roundabouts & Accidents due to improper yielding or lane changes in roundabouts. \\
        7 & Crash at Ramp Mergers & Merging errors or failure to yield causing highway on-ramp accidents. \\
        8 & Jaywalking Pedestrians & Pedestrians crossing outside crosswalks, demanding driver alertness. \\
        9 & Lane Shifting Behavior & Evaluating the driver’s timing and judgment in shifting lanes on highways. \\
        10 & Compliance to Yellow Light & Testing driver response to a sudden yellow light while approaching. \\
        11 & Slow Car \hspace{5mm} Encounter & Encountering a slow vehicle ahead, testing decision-making and patience. \\
        12 & Crash at Zipper Lane Merge & Evaluating lane merge skills and attentiveness to road signs. \\
        \bottomrule
    \end{tabular}
    \vspace*{-2em}
\end{table}

\subsection{Driving Style Data}

Before taking the VR user study, each participant is asked to fill out a questionnaire that encompasses 50 items to analyze the driving style of the participant based on the Multidimensional Driving Style Inventory (MDSI) \cite{TAUBMANBENARI2016179} questionnaire. The MDSI is a pivotal tool for assessing diverse driving behaviors. The MDSI distinctly differentiates between "driving skills" defined as a driver's capacity to control the vehicle and adeptly handle complex traffic situations and "driving style," which encompasses the habitual patterns of behavior demonstrated in these scenarios. This distinction facilitates a comprehensive analysis of the interplay between driving skills and behavioral tendencies. The MDSI categorizes participants into eight distinct driving styles based on their responses: (a) dissociative; (b) anxious; (c) risky; (d) angry; (e) high velocity; (f) distress reduction; (g) patient; (h) careful. The following are sample questions from each category of the MDSI questionnaire, rated on a Likert scale of 1 to 6 from ``less likely'' to ``highly likely'', for analyzing individual driving styles:

\begin{itemize}
\vspace*{-0.25em}
    \item I nearly hit something due to misjudging my gap in a parking lot (Dissociative)
    \item It worries me when driving in bad weather (Anxious)
    \item I enjoy the excitement of dangerous driving (Risky)
    \item I blow my horn or “flash” the car in front as a way of expressing frustrations (Angry)
    \item In a traffic jam, I think about ways to get through the traffic faster (High Velocity)
    \item I meditate while driving (Distress Reduction)
    \item When a traffic light turns green and the car in front of me doesn’t get going, I just wait for a while until it moves (Patient)
    \item I am always ready to react to unexpected maneuvers by other drivers (Careful)
    \vspace*{-0.25em}
\end{itemize}
\noindent
where the information in parenthesis denotes the driving style category to which the item belongs.

After collecting data from the MDSI questionnaires by all the participants, the responses are processed using individual weights for each question as specified by the authors. This process yields an eight-dimensional vector representing the participant's driving style. This vector serves as a ground truth label for training a machine learning model. The model aims to predict the user's driving style based on various driving metrics, including trajectory, speed, acceleration, magnitude, lane change count, and even jerk values. \\

\vspace{-1.5em}
\subsection{Sensory Data}

After completion of the MDSI questionnaire, the participants move on to the VR component of the user study. The sensory data is simultaneously collected using SUMO's {\em Floating Car Data} (FCD) output format in our traffic simulator and a few other data through the game engine. \\
\vspace{-2mm}

\subsubsection{FCD Output Data from SUMO}



The Floating Car Data (FCD) output format derived from the Simulation of Urban MObility (SUMO) traffic simulator offers detailed records of diverse parameters for each vehicle within the simulated environment. These parameters encompass precise {\em positional coordinates} in the x, y, and z axes, {\em vehicle velocity}, {\em steering angle}, and {\em lane index} at every discrete time step. Serving as a highly accurate surrogate for a GPS, this dataset substantially augments the granularity and precision of data acquisition processes, thereby facilitating a comprehensive comprehension of individual vehicle dynamics.

Moreover, the FCD output permits the computation of supplementary parameters such as acceleration, deceleration, magnitude, alterations in steering angle, and tally of lane changes. These derived metrics contribute to the enhancement of predictive capabilities within modeling frameworks. Additionally, the analysis of jerk, a derivative of acceleration, further enriches the dataset, enabling in-depth investigations into vehicle behaviors and traffic flow dynamics. \\

\vspace{-2mm}

\subsubsection{Logged Data from Unity}

In addition to the Floating Car Data (FCD) acquired from our traffic simulator, supplementary data has been gathered directly from the game engine, focusing primarily on the ego vehicle and driver-related information. This supplementary dataset includes records of the {\em ego agent's translation and rotation} in the X, Y, and Z dimensions, {\em vehicle speed}, {\em brake force}, and {\em steering angle}, logged using a Logitech steering wheel input. Additionally, {\em scenario-specific details} such as the positions of deer and pedestrians are logged within their respective scenarios. To further enrich the collected data and facilitate the analysis of driving behavior, the {\em eye movement data of the participants} is tracked and logged using the Meta Quest Pro device, capturing metrics such as eye position, gaze direction, and precise eye gaze location. The integration of eye-tracking data enables the assessment of the driver's attentiveness and focus level, serving as a validation metric for subsequent analysis.

Overall, the sensor data collected from both the traffic simulator and the game engine are essential components in evaluating the correlation between driving styles derived from the Multidimensional Driving Style Inventory (MDSI) questionnaire and observed driving behaviors within the virtual reality (VR) simulator. This comprehensive dataset facilitates the computation of correlations between self-reported driving styles and real-world driving actions, thereby enhancing our understanding of human driving behavior in simulated environments. Furthermore, leveraging this dataset enables the training of motion forecasting models by incorporating contextual information from the scenario enhancing safety and reliability in autonomous vehicle operations, particularly in scenarios involving mixed levels of autonomy. Thus, this approach not only provides valuable insights into human driving behavior but also offers practical solutions for improving the performance of autonomous vehicles across diverse driving environments.

\section{Experiments and Validation}

The dataset comprises approximately 1,205 total trajectories collected from 110 participants. The comparison between self-classified driving styles and the derived top two driving styles among participants reveals significant insights into the accuracy of the questionnaire methodology. With {\bf 83.2\%} participants showing at least one matching style between their self-classification and the derived styles, it suggests that the questionnaire effectively captures participants' self-perceived driving behaviors. This high matching rate also indicates a reasonable self-awareness among participants and validates the methodology used to derive these driving styles. Conversely, the 16.8\% participants with little to no matched styles suggest potential discrepancies, likely due to differences in self-perception, limitations in the questionnaire design, or the inherent complexity of driving behaviors. These findings underscore the questionnaire's overall effectiveness, while also pointing to areas for potential refinement to better capture the full spectrum of driving styles. To {\bf validate the significance of DISC, we analyze it under three aspects}: (a) Correlation Analysis; (b) Influence of Factors on Driving Style; and (c) Sensory Data Measurement Analysis. These analyses are detailed below. 

\subsection{Correlation Analysis}

In this section, we highlight the significance of our dataset by analyzing the correlation between driving personality traits, as measured by the Multidimensional Driving Style Inventory (MDSI), and actual driving behavior within the virtual simulator. Both Pearson (linear) and Spearman (monotonic) correlation coefficients are calculated between various driving data parameters and the driving style vector for each participant across different scenarios. This approach allows us to assess whether a linear or monotonic relationship exists, reflecting real-world driving patterns.

\begin{table*}[t]
\small
\vspace*{-1.25em}
  \caption{Pearson and Spearman correlation table between driving style components from MDSI and sensor measurements for Scenario 1 - Sudden lane intercept. Pearson correlation assesses the linear relationship between two continuous variables, while Spearman correlation evaluates the monotonic relationship, regardless of linearity.}
  \label{CorrelationAnalysis}
  \centering
    \begin{tabular}{l|c|ccccccc}
      \toprule
      \multirow{2}{*}{\textbf{DRIVING STYLE}} & \multicolumn{7}{c}{\textbf{SENSOR MEASUREMENTS}} \\
      \cmidrule(lr){2-8}
       & {\text Coefficient} & $\Sigma \text{Magnitude}$ & $\Sigma \text{Acceleration}$ & $\Sigma \text{Speed}$ & $\Sigma \text{Steering Angle}$ & $\Sigma \text{Lane}$ & $\Sigma \text{Jerk}$ \\
      \midrule
      \multirow{2}{*}{\textbf{Dissociative}} & $r_{\text{Pearson}}$ & \textbf{0.3322} & -0.0344 & \textbf{0.4059} & 0.0154 & 0.0152 & -0.0001 \\
      & $r_{Spearman}$ & \textbf{0.4791} & 0.0401 & \textbf{0.4953} & 0.0338 & 0.0153 & 0.0215 \\
      \midrule
      \multirow{2}{*}{\textbf{Anxious}} & $r_{\text{Pearson}}$ & 0.0578 & -0.0089 & 0.0719 & 0.0153 & 0.0028 & 0.0020 \\
      & $r_{\text{Spearman}}$ & 0.0790 & -0.0070 & 0.0815 & 0.0231 & 0.0048 & 0.0068 \\
      \midrule
      \multirow{2}{*}{\textbf{Risky}}  & $r_{\text{Pearson}}$ & 0.1078 & -0.0020 & 0.1300 & 0.0124 & 0.0079 & -0.0057\\
      & $r_{\text{Spearman}}$ & 0.0938 & 0.0268 & 0.0964 & 0.0057 & 0.0061 & 0.0085 \\
      \midrule
      \multirow{2}{*}{\textbf{Angry}}  & $r_{\text{Pearson}}$ & -0.2769 & 0.0248 & {\bf -0.3390} & 0.0026 & -0.0071 & 0.0007 \\
      & $r_{\text{Spearman}}$ & -0.3742 & -0.0213 & {\bf -0.3873} & -0.0241 & -0.0080 & -0.0180 \\
      \midrule
      \multirow{2}{*}{\textbf{High Velocity}} & $r_{\text{Pearson}}$ & \textbf{0.3331} & -0.0333 & \textbf{0.4074} & 0.0195 & 0.0135 & -0.0042 \\
      & $r_{\text{Spearman}}$ & \textbf{0.4685} & 0.0520 & \textbf{0.4848} & 0.0384 & 0.0145 & 0.0227 \\
      \midrule
      \multirow{2}{*}{\textbf{Distress Reduction}} & $r_{\text{Pearson}}$ & \textbf{-0.3417} & 0.0367 & \textbf{-0.4184} & -0.0160 & -0.0131 & 0.0004 \\
      & $r_{\text{Spearman}}$ & \textbf{-0.4475} & -0.0481 & \textbf{-0.4636} & -0.0426 & -0.0138 & -0.0262 \\
      \midrule
      \multirow{2}{*}{\textbf{Patient}} & $r_{\text{Pearson}}$ & 0.0986 & -0.0126 & 0.1217 & -0.0177 & -0.0056 & 0.0037 \\
      & $r_{\text{Spearman}}$ & 0.1512 & 0.0095 & 0.1556 & 0.0011 & -0.0050 & 0.0072 \\
      \midrule
      \multirow{2}{*}{\textbf{Careful}} & $r_{\text{Pearson}}$ & -0.1248 & 0.0100 & -0.1516 & -0.0230 & -0.0056 & 0.0038 \\
      & $r_{\text{Spearman}}$ & -0.1069 & -0.0221 & -0.1102 & -0.0105 & -0.0058 & -0.0062 \\
      \bottomrule
    \end{tabular}%
  \vspace*{-1.25em}
\end{table*}

Table \ref{CorrelationAnalysis} presents six notable correlations using {\bf Pearson coefficients}, emphasized in bold, demonstrating that {\em the specific associations between driving style assessments and observed behaviors are robustly linear}. Similarly, {\bf Spearman coefficients} highlight {\em monotonic relationships, suggesting a consistent association between variables, even when not strictly linear}. These findings reveal strong associations between driving behaviors and specific driving styles, which align with real-world human driving patterns, further validating the dataset's relevance and statistical significance.

For example, in Scenario 1, where a vehicle suddenly merges into our lane, we observe a strong {\em positive linear correlation} between `High-Velocity' drivers, `Dissociative' drivers, and Speed and Magnitude parameters, indicating faster speeds and larger movements, potentially reflective of a more impulsive response. In contrast, there is a {\em strong negative correlation} between the distress driving style and speed, indicating that 'distress-reduction' drivers tend to drive slower under specific conditions. These findings suggest that for modeling high-velocity drivers, focusing on metrics such as speed and magnitude with a linear model is effective. These correlations reflect typical human driving behaviors in real-world accident scenarios, enhancing the realism and applicability of the dataset for studying driver responses. For further details on the statistical significance of the results, please refer to the accompanying website.

The ability of our dataset to accurately capture and correlate driving behaviors with sensor measurements underscores its value as a tool for predictive modeling. By integrating MDSI-derived driving personalities with empirical data from the simulator, we enhance our ability to understand and predict human driving behavior, particularly in mixed-autonomy environments. This understanding is pivotal for autonomous driving technologies and formulating effective traffic safety strategies.

\vspace{-0.15em}

\subsection{Influence of Various Factors on Driving Style}

\noindent
Considering the top 2 driving styles provides a more nuanced understanding of driving behaviors, acknowledging the complexity of human behavior. This approach helps identify contradictory traits and offers insights into {\em how different personality factors interact}, allowing for more tailored interventions and strategies. Analyzing the top 2 driving styles reveals diverse personality combinations. Expected pairs, like "Careful and Patient," indicate safe driving, while contradictory pairs, such as "Patient and Risky" or "Angry and Careful," highlight the complex interplay of personality traits and their impact on driving habits.





\begin{figure*}[t]
\vspace*{-1em}
    \centering
    \includegraphics[width=0.85\textwidth, height=7.75cm]{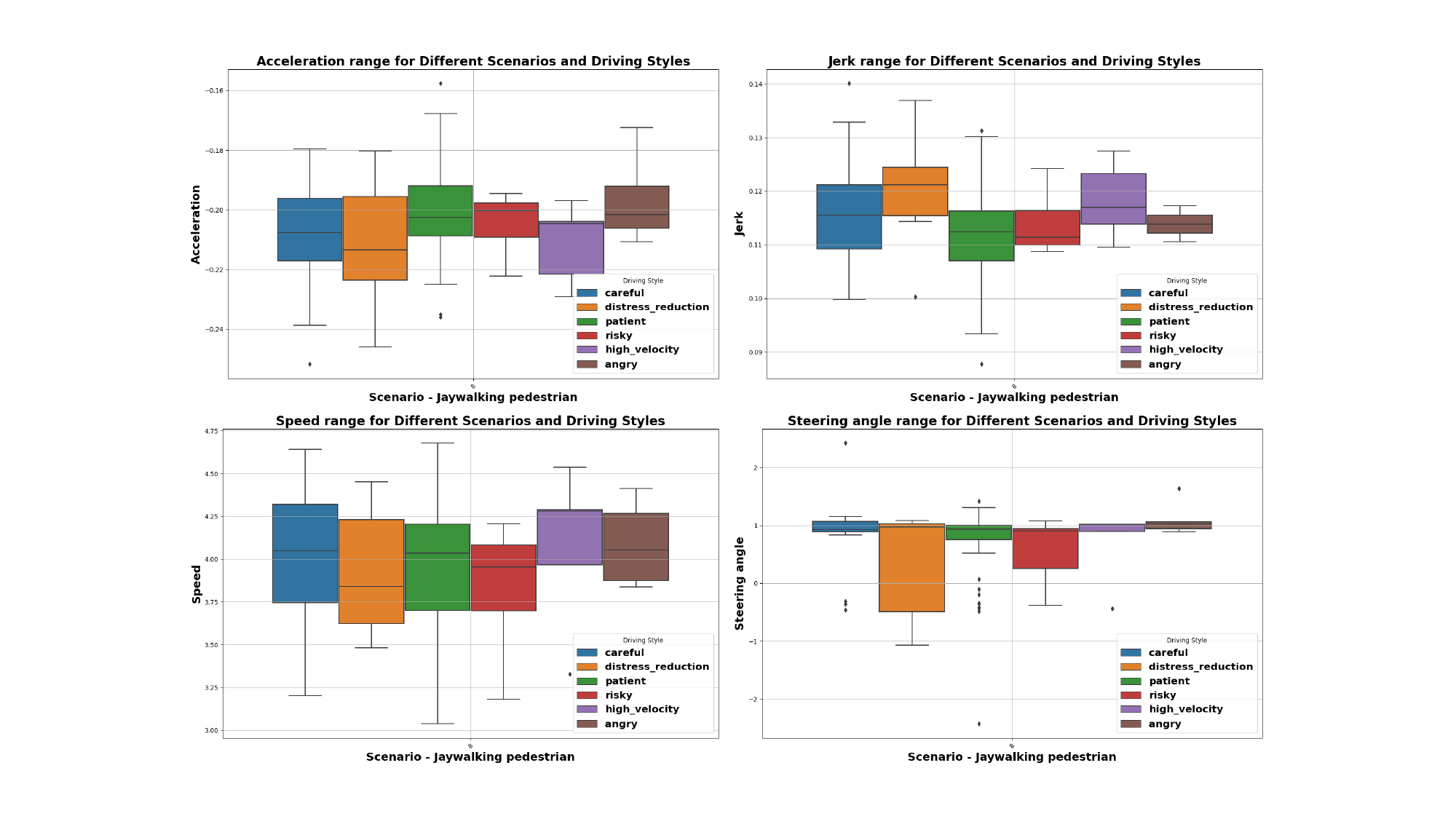}
    \vspace*{-1em}
    \caption{ \textbf{Plots of sensory values:} The figure displays plots of acceleration, jerk, speed, and steering angle, representing distinct driving styles amidst jaywalking pedestrian scenarios, arranged from the top right to the bottom left. Observations reveal diverse patterns in sensory dynamics, providing insights into the driving behavior of drivers encountering pedestrians and enabling driving personality-conditioned trajectory prediction.}
    \label{fig:boxplots}
    \vspace*{-1.5em}
\end{figure*}

\subsection{Sensory measurements Analysis}

Figure~\ref{fig:boxplots} depicts the range of sensory readings within a specific vehicular context involving an unexpected pedestrian presence for each driving style. Notably, acceleration averages for {\em high-velocity} and {\em angry} driving modalities manifest lower values, indicative of less frequent accelerative actions and a constrained range. Conversely, alternative driving styles exhibit elevated acceleration levels, suggesting heightened frequency and broader dispersion.

Additionally, jerk, the temporal derivative of acceleration, exhibits divergent patterns across driving modalities. Instances of {\em distress reduction} driving illustrate heightened jerks, mirroring the heightened emotional states of these operators. This phenomenon is echoed in steering wheel angle data, where distress reduction driving showcases amplified values, indicative of increased stress responses during pedestrian encounters.
Furthermore, speed metrics within the high-velocity driving style range register diminutive values, in concordance with the driving proclivities of these individuals.

\subsection{Qualitative Observations During the VR Driving Study}

VR driving study observations revealed some key insights:

\begin{itemize}
    \item Skilled drivers navigated accident scenarios effectively by adhering to traffic rules, maintaining safe distances, and staying alert to their surroundings.

    \item One participant successfully avoided a potential collision with a deer by anticipating its crossing upon entering a forested area, indicating heightened environmental awareness.


    \item The high level of immersion in the simulator occasionally caused participants to miss facilitator instructions, highlighting the simulator's effectiveness in creating realistic scenarios.

    \item Participants with international driving experience often ignored specific traffic signs, such as yield and stop signs, reflecting variations in global driving norms.
\end{itemize}

\subsection{Common Trajectory Dataset Format}


The dataset can be standardized using the ScenarioNet\cite{li2023scenarionetopensourceplatformlargescale} format, facilitating integration with well-established datasets such as Waymo, nuScenes, and nuPlan, all of which have specific conversion methods to this format. Notably, the Argoverse2 dataset can also be converted to ScenarioNet through the UniTraj\cite{feng2024unitrajunifiedframeworkscalable} framework. Additionally, the DISC dataset, which is converted in ScenarioNet format, can be combined with these larger datasets to enhance trajectory prediction models by adding accident scenarios and diverse driving behaviors.

UniTraj plays a crucial role in unifying datasets by aligning them to a uniform time interval of 8 seconds, and the DISC dataset is also converted into the same time interval, making it suitable for integration with other large-scale datasets. Furthermore, road maps associated with these datasets can also be converted into ScenarioNet format, ensuring consistent representation across different data sources.

Leveraging the ScenarioNet format allows the DISC dataset, to be easily integrated and fully utilized within the UniTraj framework, which is recognized as a SOTA trajectory prediction model for the nuScenes. A potential application would be the addition of a driving style prediction model to UniTraj's prediction pipeline, allowing for joint predictions of trajectories \& driving styles. UniTraj offers built-in models such as MTR\cite{shi2023motiontransformerglobalintention}, AutoBot\cite{girgis2022latentvariablesequentialset}, and Wayformer\cite{nayakanti2022wayformermotionforecastingsimple}, with MTR being the state-of-the-art model for the Waymo Dataset. By converting the DISC dataset into ScenarioNet format, the combined benefits of UniTraj, MTR, and ScenarioNet can be fully exploited to advance research in joint trajectory and driving behavior prediction.

\subsection{Limitations}
\vspace*{-0.25em} The size of DISC is limited, but it can be easily combined with other real-world datasets and further enhance scenario diversity otherwise challenging to capture. The VR simulator cannot fully replicate real-world driving, including physical sensations and environmental variations.  Nevertheless, DISC captures the most important features, such as trajectory data, vehicles, pedestrians, and the road network system states as a whole, adequate for training autonomous vehicles.  Drivers may exhibit riskier behavior in VR, and technological constraints such as frame rates, graphical fidelity, and latency may affect data quality. Additionally, motion sickness in some participants may influence their performance. Like poorer image quality, they can be addressed using adversarial training to improve robustness in future research.

\section{CONCLUSION}
In this paper, we have introduced the first dataset, {\em\textbf{DISC}}, designed to analyze human driving behavior to advance research on autonomous vehicles in mixed autonomy. The DISC dataset fills a critical gap by providing detailed data on different driving behaviors under pre-crash conditions, essential for improving autonomous vehicle safety. One such application is to predict trajectories of nearby vehicles during pre-crash scenarios, based on observed driving behaviors so far and making safer decisions based on the predictions. To achieve this, we have leveraged an Extended Reality (XR) environment using an immersive display and controllers, where participants are immersed in driving through 12 pre-crash scenarios and adverse weather. Our experiments demonstrated a strong correlation between the collected data and driving styles, indicating that the DISC dataset can effectively assess driving behaviors. DISC can be further used to recreate driving scenarios using the collected data and generate complex output, such as video footage and LiDAR data. These advances can notably improve safety of mixed autonomy, leading to safer driving for all on the road.






\section*{ACKNOWLEDGMENT}

This research is supported in part by Barry Mersky \& Capital One E-Nnovate Endowed Professors \& ARO DURIP Grant.



\vspace{15mm}
\bibliographystyle{ieeetr}

\end{document}